\documentclass{article}
\usepackage{spconf,amsmath,graphicx}
\usepackage{cite}
\usepackage{microtype}
\usepackage{amssymb}
\usepackage{url}
\usepackage[hidelinks]{hyperref}

\def\R{{\mathbb R}}
\def\C{{\mathbb C}}

\DeclareMathOperator*{\argmin}{argmin}

\title{Structured Dropout for Weak Label and Multi-Instance Learning and Its Application to Score-Informed Source Separation}
\name{Sebastian Ewert, Mark B. Sandler\thanks{This work was funded by EPSRC grant EP/L019981/1.}}
\address{Queen Mary University of London, UK}
\begin{document}
\ninept

\maketitle

\begin{abstract}
Many success stories involving deep neural networks are instances of supervised learning, where available labels power gradient-based learning methods. Creating such labels, however, can be expensive and thus there is increasing interest in weak labels which only provide coarse information, with uncertainty regarding time, location or value. Using such labels often leads to considerable challenges for the learning process. Current methods for weak-label training often employ standard supervised approaches that additionally reassign or prune labels during the learning process. The information gain, however, is often limited as only the importance of labels where the network already yields reasonable results is boosted. We propose treating weak-label training as an unsupervised problem and use the labels to guide the representation learning to induce structure. To this end, we propose two autoencoder extensions: class activity penalties and structured dropout. We demonstrate the capabilities of our approach in the context of score-informed source separation of music.
\end{abstract}
%

\section{Introduction}
\label{sec:intro}

Training neural networks is often a complex endeavor, which is somewhat simplified if clear labels are available in a supervised learning scenario \cite{LeCunBH2015_DeepLearning_NATURE}.
Annotating data, however, is typically an expensive process and hence there is increasing interest in using unlabeled data to support learning using smaller labeled dataset (semi-supervised learning) 
\cite{HintonOT2006_UnsupervisedPretrainDBN_NC}.
Another approach to address the cost issue are \emph{weak labels}, where the annotator provides only relatively coarse information. 
For example, an annotator might indicate whether a car is shown in a 5-minute video, without specifying when, where or how long it is visible. Settings based on this type of weak labels are often referred to as \emph{multiple-instance learning}, as one label provides spatially or temporally coarse information for a bag of instances \cite{KeelerRL1990_IntegratedSegmentationRecognition_BOOK}.
Another type of weak label specifies several possible outcomes: A picture shows either a bird, a plane or a dark cloud, or in regression a value might be somewhere in an interval.
For these two types, a weak label provides rough information about the existence of a concept, while the lack of a weak label gives clear information about its non-existence.

The effort to create weak labels is often only a fraction of highly detailed annotation.
For learning, however, weak labels present considerable challenges \cite{FouldsF2010_ReviewMultiInstance_TKER}.
For example, training a frame-wise car-detector can be difficult using weak labels if in a 5-minute clip labelled with `car' the car is actually only shown for 3 seconds.
A first naive way is to use supervised learning methods using the label `car' for each frame in the 5-minutes block. 
Multiple-instance learning methods try to improve on this concept by iteratively re-labeling instances as negative examples in a block \cite{AndrewsTH2002_SVMForMIL_NIPS} or pruning them \cite{ChenBW2006_MILESMultipleinstance_TPAMI,HouSKGD2015_PatchBasedDetection_ARXIV}
if they are not clearly detected as positive after an initial naive training step, see also \cite{MandelE2008_MultipleInstanceLearningMIR_ISMIR} for a comparison.
Alternatively, one can learn using only the frame which yielded the clearest positive response in a block \cite{ZhouZ2002_NeuralNetworksMultiInst_ICIIT}.
As a special case, one can use so called saliency maps to improve the temporal accuracy for networks with multi-frame inputs \cite{Schlueter2016_WeakLabelsSingingDetect_ISMIR}.
However, often such approaches are not effective \cite{Schlueter2016_WeakLabelsSingingDetect_ISMIR}, as re-labeling/pruning methods only re-interpret examples that are already correctly classified by the network (limiting the gain in information for the learning process), while highest-saliency learning ignores a considerable amount of annotated data that could be potentially useful for training.

All of these measures remain supervised learning methods, which requiring clear output labels cannot directly account for the fact that weak labels have an inherent uncertainty. In this paper, we propose to treat learning based on weak labels fundamentally as an unsupervised learning problem and use the labels only as a guidance to encourage structure in the learned representations.
More precisely, we start from the standard autoencoder architecture \cite{HintonS2006_ReducingDimensionalityData_Science}, where one network transforms the input to a low-dimensional representation, which is then used to re-synthesize the input -- differences between the input and output are measured using a \emph{reconstruction error} term.
Usually, this learned representation is not readily interpretable, which we change using the weak labels.
To this end, before training, we associate each class to be detected with a number of units in the representation layer.
Then, using the class activity information provided by weak labels, we encourage using an \emph{activity cost} term on the representation that the units corresponding to the inactive classes are zero.
This way, the encoder must model the input using only the units for the active classes in each frame.
After training, activity in the output of this structured representation layer is already a surprisingly good indicator for class activity which only needs to be refined further in standard end-to-end learning (either as improved input or as a new, more accurate output target).

To enforce these activity constraints, one typically needs to drastically favor the activity cost over the reconstruction error.
However, this sometimes had detrimental effects on the training efficiency: the gradient is dominated by the activity cost in this case and thus gradient-based learning methods often set all network weights to zero, thus deactivating all activity (i.e. ignoring the reconstruction error). This often slowed down training considerably.
Therefore, we propose a second extension to accelerate training. More precisely, during a first training stage we propagate the input through the network and force-set the units associated with inactive classes to zero. This simple idea can be interpreted in several ways. First, it is a variant of the well-known stochastic dropout technique for regularization \cite{SrivastavaHKSS2014_Dropouty_JMLR}, just that our dropout is deterministic and induces structure instead of noise. Second, the decoder part of the network can operate early during training under conditions it will find once the network is fully trained, resembling properties of batch-normalization \cite{IoffeS2015_BatchNormalization_ARXIV}. Third, the gradient is `sharpened' crossing the representation layer and affects in each input frame particularly those parts of the encoder network that are responsible for creating the active class representations. 

As a further interpretation, this procedure can be seen as a non-linear extension of \emph{non-negative matrix factorization (NMF)} in combination with certain activity constraints \cite{EwertM12_ScoreInformedNMF_ICASSP}. This leads to an intuitive interpretation of our approach as a part-based representation of the input, similar to NMF. Therefore, we motivate our procedure starting from the well-known NMF and demonstrate the capabilities of our method using a task previously addressed with NMF: Score-informed source separation of music signals \cite{EwertM12_ScoreInformedNMF_ICASSP}. In this case, we have weak labels providing coarse information about the activity of notes, without specifying intensity or progression parameters. As shown by our experiments, our proposed system surpasses the NMF results by more than 0.5dB SDR without any specific hyperparameter tuning.

The remainder is organized as follows. In Section~\ref{sec:proposedMethod} we describe our proposed method in more detail. In Section~\ref{sec:applSISS}, we show how our method can be employed for score-informed source separation and discuss in Section~\ref{sec:experiments} our experimental results. We conclude in Section~\ref{sec:conclusions} with a prospect on future work.

\section{From NMF to An Autoencoder with Structured Dropout}
\label{sec:proposedMethod}

The basic idea behind non-negative matrix factorization is to represent a series of $N$ input vectors in the matrix $V \in \R_{\ge 0}^{M \times N}$ as a product of two matrices $W \in \R_{\ge 0}^{M \times K}$ and $H \in \R_{\ge 0}^{K \times N}$ with $K<M$.
If we associate a specific class to be detected with one of the $K$ NMF components (or a group thereof), we can incorporate information from weak labels into the NMF learning process by setting entries in $H$ to zero: e.g. to express that class $k$ will be inactive in input $n$, we can set $H_{k,n}$ to zero -- using multiplicative rules to iteratively update $W$ and $H$ as usually done, these constraints will remain active throughout the learning process \cite{RaczynskiOS07_HarmonicNMF_ISMIR}. This way, we only specify which classes are not active but do not specify the exact intensity a class should have or whether it should be active at all.

To translate this NMF-based approach to incorporating weak labels to the world of neural networks, we first see that NMF and its learning process resemble the one of an autoencoder quite closely. Assuming we obtained a $W$ and $H$ with $V \approx WH$ using NMF, it follows that $H \approx W^{+}V$, where $W^{+} \in \R^{K \times M}$ is an approximation to a left-inverse of $W$, for example the Moore-Penrose pseudoinverse of $W$ or simply the transpose of $W$ if its columns are approximately orthogonal. We obtain $V \approx WW^{+}V$, which is an autoencoder with linear activation functions and some non-negativity constraints on its weight matrices \cite{HintonS2006_ReducingDimensionalityData_Science}.

\begin{figure}
\centering
\includegraphics[width=8cm]{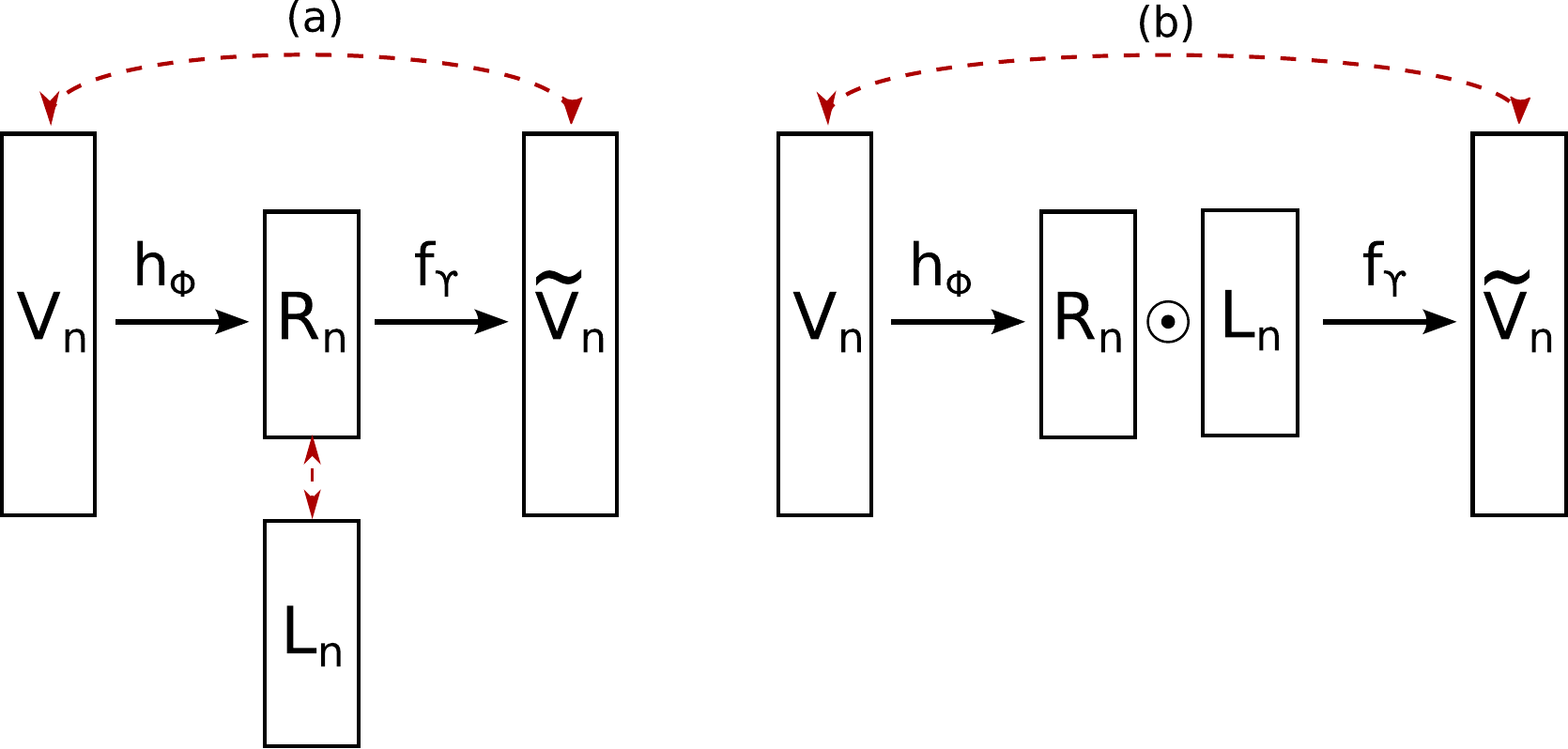}
\caption{Incorporating weak labels into an autoencoder (red connections indicate distance measures): (a) activity cost approach, (b) structured dropout approach.}
\label{fig:introFig}
\end{figure}

Let us generalize this a bit further. In particular, let $h_{\Phi}: \R^M \to \R^K$ and $f_{\Upsilon}: \R^K \to \R^M$ be multi-layer (feed-forward) networks with corresponding weight parameters in $\Phi$ and $\Upsilon$, respectively. From an autoencoder point of view, $h$ is the encoder or analysis part transforming the input $V_n$ to a $K$ dimensional representation $R_n := h_{\Phi}(V_n)$, that is then used by the decoder or synthesis part $f$ to reconstruct the input from the low-dimensional information, compare Fig.~\ref{fig:introFig}a. Here, $V_n$ denotes the $n$-th column of $V$. From an NMF point of view, $h$ and $f$ generalize the matrix products $W^{+}V$ (as above) and $WH$ to more general non-linear functions.
The autoencoder $f_{\Upsilon} \circ h_{\Phi}$ is usually trained minimizing a distance function comparing the input with the reconstruction:
\begin{equation}
\argmin_{\Upsilon, \Phi} c(\Phi, \Upsilon, V) = \frac{1}{N} \sum_{n=1}^N{d(f_{\Upsilon}(h_{\Phi}(V_n)) , V_n)},
\end{equation}
where $d$ is a distance or divergence.
This type of unsupervised feature-learning has been found useful in a variety of tasks -- however, the resulting low dimensional representation usually cannot easily be interpreted, in particular if $f$ and $h$ represent deep networks \cite{HintonS2006_ReducingDimensionalityData_Science}.

Next, we extend this standard autoencoder to impose structure on the output of $h$. However, we cannot directly use the same approach as in NMF as we do not employ optimization rules based on multiplicative updates. Therefore, we propose a different approach and train the autoencoder to yield a network that behaves similarly. To this end, we first associate each unit (or dimension) in the output of $h_{\Phi}$ with a specific class used in our weak labels.
Note that each class can (and often will) be associated with several units -- in general, this is part of the application specific design process. 
Optionally, we can add some free units useful for noise or general background modelling. Next, we exploit the availability of weak labels to create a matrix $L \in \{0,1\}^{K \times N}$, which encodes the potential class activity in each input frame. 
More precisely, we set $L_{k,n}=1$ if there is a weak label specifying that the class associated with unit $k$ is potentially active in input frame $n$, while free units are always set to one. This approach generalizes ideas used in \cite{StowellT2015_DenoisingPartAutoencoder_ARXIV}, which used a single class: noise or no-noise. We can now incorporate $L$ into our objective function to discourage activity in units that should not be active in a frame:
\begin{multline}
\argmin_{\Upsilon, \Phi} c(\Phi, \Upsilon, V, L) = \frac{1}{N} \sum_{n=1}^{N}d(f_{\Upsilon}(h_{\Phi}(V_n)) , V_n) \\
+ \lambda \, \big\| (1-L_n) \odot h_{\Phi}(V_n)\big\|_2^2,
\label{eq:zeroConstrCostTermOpt}
\end{multline}
where $\lambda$ is a term balancing the reconstruction error and our activity cost and $\odot$ is the Hadamard product (point-wise multiplication), see also Fig.~\ref{fig:introFig}a. Using this modified $\ell^2$ penalty on the output of $h_{\Phi}$, we penalize activity in units associated with classes known to be inactive and exclude the potentially active ones.
Further, in contrast to classical supervised methods, we do not specify what should happen if a class is active, i.e.\,what value a unit should have. However, if a unit becomes active it is often a spatially/temporally accurate indicator that a certain class is present.  

In practice, however, our new objective function led to some numerical problems, especially with deeper networks. In particular, to enforce the activity constraints, the parameter $\lambda$ needs to be set to a relatively high value. As a consequence, the gradient will be dominated by our new term, which in many cases steered the learning process towards a local minimum with all weights in $\Phi$ set to zero, rather accepting the reconstruction error than making a mistake w.r.t.\,activity costs. A possible workaround was to employ a schedule for $\lambda$ that would slowly increase its value with the number of training iterations -- setting such a schedule, however, can be difficult, as it depends on properties of the error surface described by the objective function $c$. If the schedule is too slow, the convergence rate can be very slow, and if it is too fast, the training might not converge.

We therefore propose a different learning scheme, which is inspired by the idea of dropout \cite{SrivastavaHKSS2014_Dropouty_JMLR}. Using the regular dropout, one multiplies the output of a network layer element-wise with a randomly generated binary dropout vector. The idea is to disable certain units in the network, which is thus encouraged to make the most important information redundantly available in the network \cite{WagerWL2013_DropoutAsRegularizer_NIPS,SrivastavaHKSS2014_Dropouty_JMLR}, a form of regularization. For our method, we will follow the same principle but use a deterministic and structured dropout vector. More precisely, we employ our binary label matrix $L$ as follows (see also Fig.~\ref{fig:introFig}b):
\begin{equation}
\argmin_{\Upsilon, \Phi} c(\Phi, \Upsilon, V, L) = \frac{1}{N} \sum_{n=1}^{N}d(f_{\Upsilon}(L_n \odot h_{\Phi}(V_n)) , V_n).
\label{eq:detDropoutOpt}
\end{equation}
This simple modification has several advantages.
First, similar to batch normalization \cite{IoffeS2015_BatchNormalization_ARXIV}, we can accelerate the training process by supplying the synthesis layers with the conditions we expect the analysis part to deliver once it is fully trained (i.e. the inactive classes are indeed inactive).
Further, similar to regular dropout \cite{SrivastavaHKSS2014_Dropouty_JMLR}, setting some units to zero naturally cuts the error back propagation and channels the entire error information to those network weights in the analysis part that are actually responsible for the active classes in a frame.
Overall, training the network using objective~(\ref{eq:detDropoutOpt}) in a first step followed by a refinement using objective~(\ref{eq:zeroConstrCostTermOpt}) typically led to a drastic acceleration, in our specific application scenario often reducing the number of iterations by a factor of $1000$ (however, we do not claim that this is generally the case). Further, with this two step procedure, we did not require a complicated schedule for $\lambda$ anymore.

\section{Application: Score-Informed Source Separation of Music Signals}
\label{sec:applSISS}

To demonstrate the capabilities of our proposed approach, we employ it in a specific application scenario.
To this end, we assume we are given an audio recording and a MIDI file encoding the uninterpreted score for a piece of music.
The note information given by the score provides coarse information when certain instruments and pitches are active.
However, the score does not specify exactly when and how notes are played, how they spectrally manifest or what their temporal progression is.
The MIDI events can thus be regarded as weak labels in both ways discussed in the introduction, i.e. the temporal information is coarse and target values are uncertain.

Based on these labels, we could use our proposed method as a basis for a (frame-wise) instrument or pitch detection method.
In this case, we would need to train an actual classifier based on our learned representation -- using it either as an improved input or as a temporally more accurate target.
However, instead of introducing this added complexity, we chose a task where we can use our autoencoder directly and still demonstrate the induced structure in the representation layer: score-informed source separation \cite{EwertPMP14_ScoreInformedSourceSep_IEEE-SPM}.
Here, the task is to extract those parts of the recording that correspond to a specified group of notes, for example, all notes associated with a specific instrument or a specific MIDI pitch. 
Next, we describe how a specific instance of our autoencoder can be used in this context.

\begin{figure}
\hspace{0.7cm}\includegraphics[width=7cm]{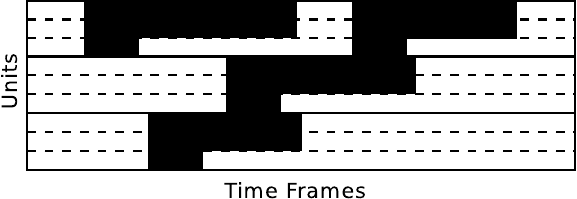}
\vspace{-0.2cm}
\caption{Example Label Matrix.}
\label{fig:exampleLabelMatrix}
\end{figure}

As a start, we use the recording and compute a short-time Fourier transform (STFT) $X \in \C^{M \times N}$ (Hann window of 93ms, stepsize 23ms) and use the frames of its magnitude $V \in \R_{\ge 0}^{M \times N}$ as input to our encoder.
To keep it simple, we use for both $f_{\Upsilon}$ and $h_{\Phi}$ standard multi-layer feed-forward networks with sigmoid activation functions (to obtain a reasonable amount of non-linearity), except for the representation and output layer where we use rectified linear activations to enforce non-negativity. For the reconstruction error, we use the generalized Kullback-Leiber divergence, which was found useful in a variety of source separation applications \cite{EwertPMP14_ScoreInformedSourceSep_IEEE-SPM}:
\[
d(a,b) := \sum_{m=1}^M{a_m \log{\frac{a_m + \varepsilon}{b_m + \varepsilon} - a_m + b_m}},
\]
with some $a,b \in \R_{\ge 0}^{M}$ and $0<\varepsilon\ll 1$.

As in \cite{EwertM12_ScoreInformedNMF_ICASSP}, we improve the temporal accuracy of the score information by 
employing the method described in \cite{EwertMG09_HighResAudioSync_ICASSP} to temporally align the MIDI file to the given audio recording.
This way, we at least roughly know where notes are being played by which instrument and which pitch they roughly have -- all further details, however, are unknown. Next, we associate each combination of instrument and MIDI pitch (corresponding to a class) used in the piece with $P\ge 2$ units in our representation layer. Each instrument-pitch class is further subdivided following ideas presented in \cite{EwertM12_ScoreInformedNMF_ICASSP}:
In each block of $P$ units, we associate the first unit with the onset of a note being `active' and the remaining units with the sustain phase being `active'.
Using these associations, we can create the label indicator matrix $L$, see Fig.~\ref{fig:exampleLabelMatrix} for an example.
More precisely, for each MIDI note we read the start and end time, the instrument-ID and MIDI pitch. 
Using the instrument and pitch information, we identify the block of $P$ associated units.
If unit $p$ in this block is the unit associated with the onset, we set an entry $L_{p,n}=1$ if frame~$n$ is in a close vicinity of the start time of the note (for example $\pm 0.5$sec to account for possible inaccuracies in the MIDI-audio alignment).
Similarly, if unit $p$ in this block is associated with the sustain phase, we set an entry $L_{p,n}=1$ if frame~$n$ corresponds to a time point between the start and end time of the note, subject to a similar temporal tolerance.
A simplified example is shown in Fig.~\ref{fig:exampleLabelMatrix}: three different combinations of instrument and pitch, and with $P=3$, we thus have $9$ units overall.
Using a total of four notes, we can see that the onset units are indicated by a short block of ones around the expected onset, and the remaining units encode the expected note length.

Using this $L$, we train our autoencoder using our proposed objectives (\ref{eq:detDropoutOpt}) and (\ref{eq:zeroConstrCostTermOpt}), using the audio recording corresponding to the score as input.
The actual separation is performed after convergence, where we can exploit the parts-based interpretation for our learned representation.
To this end, we choose a group of notes $\mathcal N$ we aim to keep.
Then, we set all entries in $L$ that do not correspond to these notes to zero.
Using this $L_{\mathcal N}$, the structured dropout will cancel all activity information related to unwanted notes when sending the input through the network.
As a consequence we now expect at the output only the part of the magnitude spectrum that corresponds to the notes to be kept.
We refer to this modified output as $\widetilde{V}_{\mathcal N}$.
Finally, we obtain our separation result via soft-masking (or Wiener filtering) $\frac{\widetilde{V}_{\mathcal N}}{\widetilde{V}} \odot X$, where the division is element-wise, and use an inverse STFT to obtain a time-domain signal.

As we will see in the next section this basic configuration already yields good separation results.
However, we found a few simple tricks to improve the separation quality even further, which might be useful in other contexts as well.
First, we observed that while our learned representation typically yields a parts-based representation, the subsequent synthesis function learned without any constraints can weaken this interpretation which then lowers the separation quality.
In particular, if entries in the weight matrices used in the feed-forward network $f_{\Upsilon}$ can contain negative entries, the network can eliminate some energy associated with a specific note based on the energy associated with another note.
The argument is the same used to compare Independent Component Analysis (ICA) and NMF \cite{LeeS99_LearningPartsNMF_Nature}: the autoencoder is building the output not in a purely constructive way starting from the learned representation.
Therefore, analogously to moving from ICA to NMF, we constrain in a variant of our method all weight matrices in $\Upsilon$ to be non-negative.
Note, as shown in \cite{ChorowskiZ2015_LearningUnderstandableNN_TNNLS}, that this constraint does not lower the network's theoretical capability to approximate arbitrary functions.
As a second extension, we used as input to our analysis network $h_{\Phi}$ not only a single frame of $V$ but provided the surrounding frames as well.
In this case, we select the center frame of the input as the target for our autoencoder, i.e. the output remains a single frame 

\section{Experiments}
\label{sec:experiments}

To evaluate our method, we conducted a series of experiments following the experimental setup used in \cite{EwertM12_ScoreInformedNMF_ICASSP}.
In particular, our task is to separate in piano recordings the notes played by the left hand from those played by the right hand. This task highlights, in contrast to general music source separation, our capability to separate arbitrary groups of notes, even originating from the same instrument. 
The dataset consists of four Bach pieces (mainly inventions) and six Chopin pieces (mainly preludes and mazurkas) and contains MIDI files from the Mutopia Project\footnote{\scriptsize\url{http://www.mutopiaproject.org}} for each piece.
As in \cite{EwertM12_ScoreInformedNMF_ICASSP}, we conduct our quantitative experiments using synthetic data.
To this end, we first synthesized the downloaded MIDI files using a high-quality, multi-sample wave table synthesizer (Native Instruments) to obtain corresponding audio recordings. Further, we synthesized the notes for the left and right hand separately for each piece, to obtain ground truth separation results. Each recording is 30 to 300 seconds long.

We implemented our proposed method in TensorFlow, with three layers for the analysis network $h$ and two layers for the synthesis network $f$ -- more layers lowered the separation quality for variants of our method without non-negativity constrains on the synthesis network weights and did not improve the quality otherwise. Each intermediate layer used $1500$ units -- the size of the input, output and representation layers were defined by the input data.
As optimizer, we used ADAM, a first-order method offering many features of second-order approaches, a property often useful in cases where the objective function needs to be minimized with a relatively high accuracy \cite{KingmaB2014_Adam_ARXIV}. All parameters were left at their recommended values \cite{KingmaB2014_Adam_ARXIV}, except for the step size which was decreased to $1/10$th of its default. Further, since we trained a new network for each piece, we used full-batch training. The optional non-negativity constraints for the synthesis network were implemented using intermediate projection onto the non-negative orthant. As a proof of concept, we did not use any specific hyperparameter optimization nor regular dropout or batch normalization.

Separation quality is assessed using the BSSEVAL toolkit \cite{VincentGF06_PerformanceMeasurement_IEEE-TASLP}. Note that, in contrast to \cite{EwertM12_ScoreInformedNMF_ICASSP}, we here use the \emph{Normalized Signal-to-Distortion Ratio (NSDR)}, where we subtract from the actual SDR value the SDR value we obtain using the full mixture recording as the separation result. The NSDR value represents the gain a method yields over simply using the original mixture and, in contrast to the plain SDR, accounts for energy differences between sound sources which are a major factor influencing the difficulty to obtain a separation.

\begin{figure}
\hspace{0.2cm}\includegraphics[width=8cm]{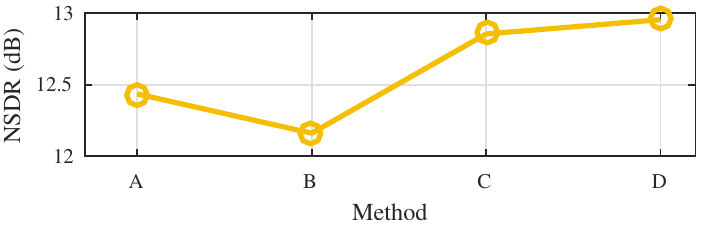}
\vspace{-0.24cm}
\caption{Averaged results for several methods.
(A): NMF baseline \cite{EwertM12_ScoreInformedNMF_ICASSP}.
(B): Prop. method.
(C): Prop. method \& non-neg. synthesis weights.
(D): Prop. method \& non-neg. synthesis weights \& multiple input frames.
}
\label{fig:results}
\end{figure}

Fig.~\ref{fig:results} shows the results for four methods. As we can see, the NMF baseline \cite{EwertM12_ScoreInformedNMF_ICASSP} already yields a relatively high separation quality of 12.4dB NSDR (method A). Using the same interpretation and dimension of the internal representation as used for NMF (i.e. $K$), our proposed method (B) without additional extensions yields results on a similar level but remains below the result for NMF.
However, if we add the non-negativity constraints on the weight matrices for the synthesis network (method C), the differences switch and our proposed method improves upon the NMF baseline by 0.4dB. These results might indicate that the pure parts-based representation might indeed be an important factor for this application. Further, in comparison to the linear NMF model, the additional layers and non-linear sigmoids might enable the autoencoder to obtain a more meaningful learned representation. Finally, in a last modification we used several consecutive frames as input to our autoencoder. This extension improved the results again by a small margin of 0.15 db NSDR, leading to an overall improvement of 0.55dB NSDR.

\section{Conclusions}
\label{sec:conclusions}

We presented a method for learning using weakly labeled data based on neural networks. Existing methods had previously tried to extend supervised learning methods to account for the often strong uncertainty in the labels. However, since these approaches typically only enhance the treatment of labels or classes for which the network already yields clear results, the impact on the learning process can be limited. As an alternative, we proposed treating the problem fundamentally as an unsupervised learning problem, i.e. starting without labels, and then only induce some structure in the resulting learned data representations based on the weak labels.
To this end, we introduced an activity cost term, which enabled us to train an autoencoder and express our uncertainty about the target value of a weak label. To accelerate the training process, we additionally proposed a structured variant of dropout, where, compared to the regular dropout, labels are used to enforce a specific structure in the network early on during training.
Our experiments based on score-informed source separation showed that indeed our proposed method can be used to induce structure in data representations learned via autoencoders.
In the future, we plan to employ our proposed method in a variety of tasks, either as a pre-processing step to provide a semantically more meaningful input representation for actual classifiers or as a training target, enhancing the label accuracy of the weak labels.

\clearpage
\bibliographystyle{IEEEbib}
\bibliography{referencesMusic}

\end{document}